\documentclass[journal=jacsat,manuscript=article]{achemso}

\usepackage[version=3]{mhchem} 

\usepackage{multicol}

\author{Tirtha Vinchurkar}
\author{Janghoon Ock}
\affiliation[ChemE]
{Department of Chemical Engineering, Carnegie Mellon University, 5000 Forbes Street, Pittsburgh, PA 15213, USA}
\author{Amir Barati Farimani}
\affiliation[MechE]
{Department of Mechanical Engineering, Carnegie Mellon University, 5000 Forbes Street, Pittsburgh, PA 15213, USA}
\email{barati@cmu.edu}

\title[An \textsf{achemso} demo]
  {Explainable Data-driven Modeling of Adsorption Energy in Heterogeneous Catalysis }

\abbreviations{IR,NMR,UV}

\begin{document}

\begin{tocentry}





\centering
\includegraphics[width=8.4cm]{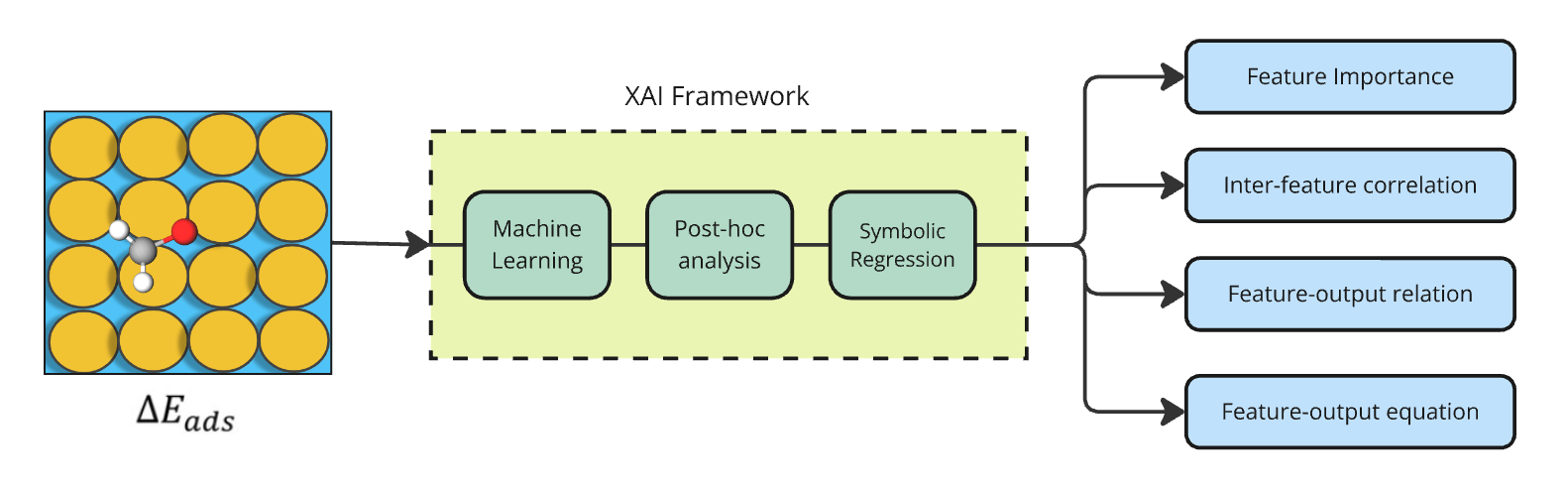} 
\label{fig:toc}
\end{tocentry}
\begin{abstract}

  
 The increasing popularity of machine learning (ML) in catalysis has spurred interest in leveraging data-driven approaches to enhance catalyst design. Despite the promise of machine learning approaches, they often function as black boxes, leaving a gap between physics-based studies and data-driven methodologies. Our study aims to bridge this gap by integrating ML techniques with eXplainable AI (XAI), specifically employing two techniques: post-hoc XAI analysis and Symbolic Regression, to unravel the correlation between adsorption energy and the properties of the adsorbate-catalyst system. Leveraging a large dataset such as the Open Catalyst 2020 (OC20) Dataset, we employ a combination of shallow ML techniques and XAI methodologies. Our investigation involves utilizing multiple shallow machine learning techniques to predict adsorption energy followed by post-hoc analysis for feature importance, inter-feature correlations, and the influence of various feature values on prediction of adsorption energy. Features such as adsorbate electronegativity, coordination number of the adsorbing atom, electronegativity of the adsorbing atom, catalyst electronegativity, and the number of atoms in the adsorbate stand out in our post-hoc analysis within our feature set for predicting adsorption energy. There is a positive correlation between catalyst and adsorbate electronegativity with the prediction of adsorption energy. Additionally, Symbolic Regression, which helps elucidate mathematical expressions describing the relationship between these important features and adsorption energy, yields results consistent with physics-based equations in previous research. It deduces a mathematical relationship indicating that coordination number of the adsorbing atom is directly proportional to the adsorption energy. Our work establishes a robust framework that integrates ML techniques with XAI, leveraging large datasets like OC20 to enhance catalyst design through model explainability.

\end{abstract} 
\section{Introduction}


Rational design of optimal catalysts remains one of the critical challenges in heterogeneous catalysis \cite{waclawek2018major}. While the primary goal is to enhance catalytic performance, the underlying physics spans across nanoscale phenomena, interfacial chemistry, and reaction engineering \cite{liu2007multi}. The multifaceted nature of catalyst surfaces and interfaces makes it particularly difficult to establish universal theoretical frameworks that can reliably predict catalyst reactivity through simple descriptors. Despite these challenges, significant progress has been made in identifying key parameters that govern catalytic behavior.

In catalyst design and development, a critical focus lies on understanding adsorption energy due to its strong correlation with catalytic reactivity, governing the interaction between reactants and catalysts \cite{norskov2002universality}. The Brønsted-Evans-Polanyi (BEP) relations illustrate volcano-shaped linear relationships between catalytic activity and adsorption energies, highlighting it as a pivotal descriptor \cite{bligaard2008heterogeneous,cheng2008bronsted,bligaard2004bronsted, gerasev2022relationship}. The BEP relation demonstrates a linear relationship between activation barrier, $E_a$, and adsorption energy, $E_{\text{ads}}$, in heterogeneous catalysis. This linear relationship implies that the catalyst's reactivity can be primarily characterized by a descriptor, $E_{\text{ads}}$. Modeling adsorption energy provides insights into reactivity and aids in catalyst screening, thus understanding the factors affecting adsorption energy is essential to identify the optimal catalyst for a target reaction.

Density Functional Theory (DFT) is a quantum chemistry-based method widely used to calculate the properties of atomic systems, including adsorption energy in catalysis. Despite its high accuracy, DFT requires substantial computational resources. Consequently, evaluating a large number of materials for catalyst design using expensive quantum chemistry-based approaches like DFT for high-throughput screening becomes infeasible. As a solution, machine learning approaches are gaining much attention due to their potential to accurately predict properties, such as energy and forces, at a reduced computational cost. Notably, extensive databases such as the OC20 dataset \cite{chanussot2021open} comprise over 1.2 million DFT relaxations of adsorbate-catalyst systems (approximately 250 million single-point calculations) across a substantially broader structure and chemistry space than previously realized. These datasets serve as the foundation for training and deploying machine learning models, aimed at predicting the energy and interatomic forces of the adsorbate-catalyst systems. The extensive availability of material databases like OC20 for adsorbate-catalyst systems has significantly accelerated the adoption of ML approaches in catalysis, particularly Graph Neural Networks (GNNs) \cite{ghanekar2022adsorbate, liao2022equiformer, gasteiger2021gemnet, xie2018crystal, schutt2017schnet, gasteiger2020directional, ock2024multimodal}.


Despite the rise of ML approaches in atomic modeling, their black-box nature obscures the underlying physical insights related to energy and force predictions. This lack of interpretability limits the further application of ML methods, as the contributing factors to predictions remain unclear. Additionally, since ML predictions inherently come with a certain level of uncertainty\cite{ock2023error}, it is more beneficial to gain insights about catalyst systems from ML modeling rather than solely relying on prediction values. To address this challenge, this paper introduces the use of XAI, which uncovers the underlying correlations behind predictions and provides human-understandable explanations, thereby enhancing domain-specific knowledge\cite{esterhuizen2022interpretable}. To achieve explainability, we implement two strategies: firstly, training shallow machine learning models such as  Adaboost Regression, XGBoost Regression, Support Vector Regression (SVR), Kernel Ridge Regression (KRR), and Least Absolute Shrinkage and Selection Operator (LASSO) Regression accompanied by post-hoc XAI analysis; secondly, applying Symbolic Regression to gain better understanding of relation between top 5 input features and adsorption energy. The former strategy highlights the importance and correlations of features, while the latter generates mathematical equations that directly expose the potential relationships between input features and target label.

\section{Results and Discussion}

\subsection{Framework}

\begin{figure*}[!hptb] 
\centering
\includegraphics[width=0.99\textwidth]{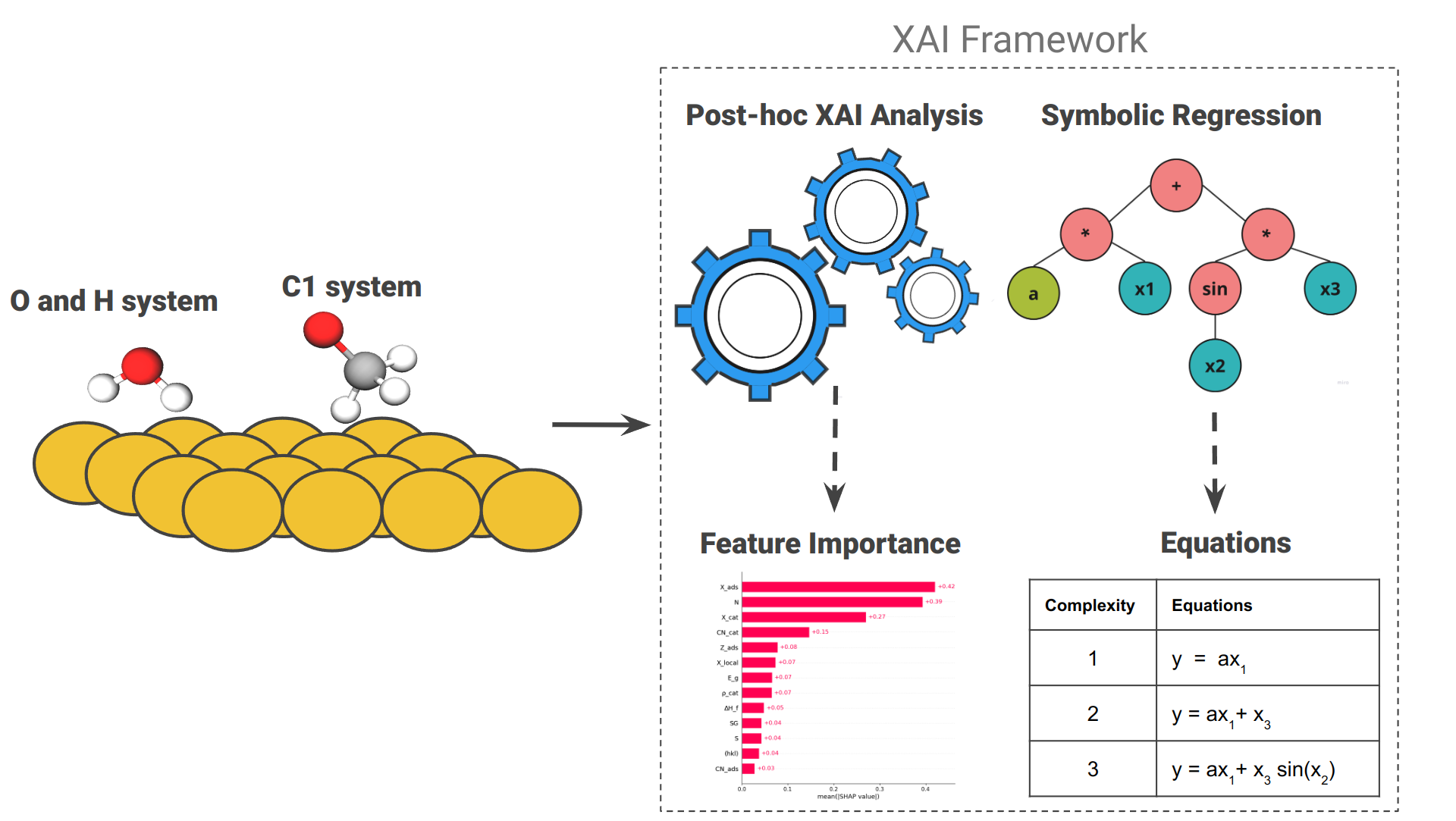} 
\caption{Overview of XAI Methods: Predicting adsorption energy with shallow machine learning models and symbolic regression. Feature importance is derived from shallow machine learning predictions through post-hoc SHAP analysis. Symbolic regression provides mathematical equations alongside its predictions.}
\label{fig:framework}
\end{figure*}

Our study leverages XAI to extract meaningful insights about the relationship between the characteristics of adsorbate-catalyst systems and the adsorption energy, which is the label value for ML predictions. We utilize XAI in two distinct ways as seen in Figure \ref{fig:framework}. First, through shallow machine learning techniques combined with post-hoc XAI analysis. The post-hoc XAI analysis is conducted using the SHAP (SHapley Additive exPlanations) library \cite{lundberg2017unified}, which provides feature importance based on Shapley values. This feature importance is crucial for understanding which characteristics most strongly influence adsorption energy, helping to guide catalyst design by highlighting key factors that impact system performance. Second, we employ Symbolic Regression to derive mathematical equations to compute the adsorption energy based on the top 5 input features with higher Shapely values. Obtaining mathematical equations to calculate adsorption energy is crucial as they enable a clear, quantitative understanding of complex relationships in data, allowing for efficient predictions, easier integration into simulations, and insights into the governing principles of adsorption phenomena. These equations are compared to equations obtained from traditional theory or experimental-based methods, highlighting their potential complementarity. 

The feature importance and correlation among features from SHAP analysis, along with the mathematical equations obtained from the Symbolic regression technique, provide better insights into the relation of adsorption energy with structural properties in a data-driven manner. The adsorption energies and feature values like electronegativity, coordination number, and so forth, for both methods are extracted from the adsorbate-catalyst systems in the OC20 dataset. The details of the feature extraction are elaborated in the following section. Integrating ML with XAI on large datasets, such as the OC20 dataset, yields valuable modeling insights that enhance our understanding of catalyst behavior. These insights can be validated through physics-based studies, providing a robust framework for informed catalyst design and accelerating advancements in the field.

\subsection{Selection of Features}

The features are carefully chosen from previous physics-based modeling research regarding adsorption energy computation. The features are categorized into three groups: adsorbate, catalyst-bulk, and surface features, as seen in Table \ref{tab:shap_values}. Each feature provides insights into electronic structure, bonding characteristics, and energetic properties of adsorption sites and catalysts. For instance, local electronegativity is determined as the geometric mean of the Pauling electronegativity of metal atoms within the first neighboring shell, offering insight into the local reactivity of catalyst surfaces\cite{li2017feature} as follows:

\begin{equation}
\label{eq:local_e}
\chi_{local} = \prod_{j=1}^{\text{1st nn}} {\chi_j^{0}}^{1/N}
\end{equation}
where $\chi_j^{0}$ is the electronegativity of atom j and N is the total number of atoms within the first neighboring shell including the adsorption site i. While, site type represented numerical classifications of different adsorption site configurations such as bridge sites, ontop sites and hollow-HPC sites.

\begin{table}[!htbp]
\centering
\caption{Input features with corresponding notations, references, and descriptions}
\label{tab:shap_values}
\resizebox{\textwidth}{!}{%
\begin{tabular}{lp{5cm}clp{5cm}}
\hline
\textbf{Category} & \textbf{Features} & \textbf{Notation} & \textbf{Reference} & \textbf{Description} \\ \hline
Adsorbate & Adsorbate electronegativity & $\chi_{ads}$ & Ishioka et al.\cite{ishioka2022designing} & Average electronegativity of all atoms in adsorbate molecule \\ 
 & Sum of adsorbate atomic numbers & $\Sigma Z_{ads}$ & Tran et al.\cite{tran2018active} & Sum of atomic numbers of all atoms in adsorbate molecule \\ 
 & Center atom coordination number & $CN_{center}$ & Calle-Vallejo et al.\cite{calle2015finding} & Coordination number of the central atom in adsorbate molecule \\ 
 & Adsorbing atom atomic number & $Z_{adatom}$ & Roy et al.\cite{roy2021machine} & Atomic number of adsorbing atom present in adsorbate \\ 
 & Adsorbing atom electronegativity & $\chi_{adatom}$ & Amos et al.\cite{amos2019feature} & Pauling’s electronegativity of adsorbing atom present in adsorbate \\ 
 & Adsorbing atom coordination number & $CN_{adatom}$ & Calle-Vallejo et al.\cite{calle2015finding} & Coordination number of adsorbing atom present in the adsorbate \\ \hline
Catalyst-bulk & Number of adsorbate atoms & $N_{ads}$ & Amos et al.\cite{amos2019feature} & Total number of atoms present in adsorbate molecule \\ 
 & Density & $\rho_{cat}$ & Li et al.\cite{li2017feature} & Density of the overall catalyst structure \\ 
 & Band gap & $E_g$ & Ma et al.\cite{ma2020machine} & Band gap of catalyst structure \\ 
 & Space group & $SG$ & Jain et al.\cite{jain2018atomic} & Space group number of catalyst \\ 
 & Formation energy & $\Delta H_f$ & Ha et al.\cite{ha2021tuning} & Formation energy of catalyst \\ 
 & Mean atomic number of bulk/catalyst & $\bar{z}_{bulk}$ & Tran et al.\cite{tran2018active} & Mean atomic number of the catalyst atoms \\ 
 & Fermi Energy & $E_F$ & Ishioka et al.\cite{ishioka2022designing} & Fermi Energy of catalyst \\ \hline
Surface & Catalyst electronegativity & $\chi_{cat}$ & Ishioka et al.\cite{ishioka2022designing} & Average electronegativity of the interacting surface atoms of catalyst \\ 
 & Effective coordination number & $CN_{cat}$ & Calle-Vallejo et al.\cite{calle2015finding} & Average coordination number of the interacting surface atoms of catalyst \\ 
 & Local electronegativity & $\chi_{local}$ & Li et al.\cite{li2017feature} & Geometric mean of the Pauling electronegativity of catalyst atoms within the first neighboring shell at the adsorption site \\ 
 & Site type & $S$ & Roy et al.\cite{roy2021machine} & Type of adsorption site based on interaction \\ 
 & Miller index & $hkl$ & Chen et al.\cite{chen2023machine} & Miller index of catalyst \\ 
 & Adsorbate-catalyst bond length & $d_{ac}$ & Ha et al.\cite{ha2021tuning} & Bond length of adsorbing atom and slab atom \\ \hline
\end{tabular}%
}
\end{table}

The features are primarily selected based on previous adsorption modeling studies, as substantiated by the references listed in Table \ref{tab:shap_values}, justifying their inclusion in ML-based adsorption energy modeling. Ishioka et al. demonstrated their machine learning models consisting of a Random Forest Classifier and Support Vector Classifier, utilized electronegativity, atomic number, and density as catalytic descriptors and could accurately predict ethylene/ethane selectivity (C2s) in the oxidative coupling of methane reaction \cite{ishioka2022designing}. The cross-validation score reached a high value of 0.844 for both models, successfully predicting three catalysts with high C2s values, which were subsequently confirmed by experimental validation. Additionally, Roy et el. showed that their model can predict adsorption energies for various intermediates accurately, leading to the identification of seven active catalysts, including \ce{CuCoNiZn}-based tetrametallic, \ce{CuNiZn}-based trimetallic, and \ce{CuCoZn}-based trimetallic alloys. The model construction considered the impact of adsorption sites, encompassing tops, bridges, and hollows, along with their neighboring atoms \cite{roy2021machine}. Calle-Vallejo et al. showcased how the utilization of coordination number, incorporating second-nearest neighbors, enabled the identification of three strategies for introducing cavity sites onto the platinum (111) surface. This enhancement proved crucial in improving its efficacy in the oxygen reduction reaction, vital for fuel cells \cite{calle2015finding}.

Moreover, Ha et al. employed formation energy as a machine learning descriptor to facilitate the discovery of high-performance single-atom catalysts for electrochemical reactions such as the hydrogen evolution reaction and oxygen evolution/reduction reactions\cite{ha2021tuning}. The values of area under the Receiver Operating Characteristic (ROC) curve ranged from 0.79 to 0.91. Furthermore, Tran et al. illustrated how atomic number serves as a key descriptor in identifying 131 \ce{CO2} reduction candidate surfaces across 54 alloys and 258 \ce{H2} evolution surfaces across 102 alloys. The Root-Mean-Squared Error (RMSE), Mean Absolute Error (MAE), and median absolute deviation for predictions were 0.46, 0.29, and 0.17 eV, respectively \cite{tran2018active}. The incorporation of structural features, notably the number of atoms alongside elemental properties, contributed to the predictive accuracy in modeling band gaps, resulting in a model capable of predicting band gaps for 2254 light-harvesting materials with a RMSE of 0.23 eV and MAE of 0.14 eV \cite{amos2019feature}. Overall, the careful selection of features, informed by both theoretical considerations and empirical evidence from the literature, provides a comprehensive framework for analyzing the complex interplay between catalyst structure, adsorption behavior, and catalytic activity.

\subsection{Post-hoc XAI analysis}

In our study, we employ SHAP analysis on shallow machine learning models to gain insights into feature importance, inter-feature correlations, and the influence of various features on predicting adsorption energy. The chosen machine learning models include Adaboost Regression with a base estimator as Random Forest Regressor, XGBoost Regression, Support Vector Regression (SVR), Kernel Ridge Regression (KRR), and Least Absolute Shrinkage and Selection Operator (LASSO) Regression. We chose the above ML models based on insights gleaned from comprehensive reviews and studies on machine learning applications for property prediction of materials to ensure the relevance of our analytical framework \cite{butler2018machine, chen2020critical}.

The dataset, comprising 24053 adsorbate-catalyst systems, is divided into an 85:15 ratio for training and testing purposes, with input features and the adsorption energy label provided to the machine learning models. Details about the dataset and data selection are elaborated in the Methods section. The performance outcomes of the models are presented in Table \ref{tab:model_performance}. The outcomes demonstrated by all models exhibit comparable performance with MAE values ranging from 0.398 to 0.617. Notably, Adaboost regression stands out as the best performer, demonstrating the lowest MAE value. Therefore, it is used as the reference model for SHAP analysis in the post-hoc analysis. This highlights an important point: post-hoc analysis heavily depends on model performance. In other words, the analysis reflects how the model interprets the features, so advancing the model is essential to gain deeper insights.

\begin{table}[!htbp]
\centering
\caption{Model Performance Comparison Based on MAE (Mean Absolute Error) in eV}
\label{tab:model_performance}
\small
\begin{tabular}{p{6 cm}cc}
\hline
\textbf{Model} & \textbf{MAE [eV]}  \\ \hline
Adaboost Regression & \textbf{0.398 ± 0.024} \\
XGBoost Regression & 0.461 ± 0.049 \\
Support Vector Regression  & 0.477 ± 0.068 \\
Kernel Ridge Regression & 0.509 ± 0.089 \\
LASSO Regression & 0.617 ± 0.065 \\ \hline
\end{tabular}
\end{table}

\begin{figure*}[ht] 
\centering
\includegraphics[width=1.0\textwidth]{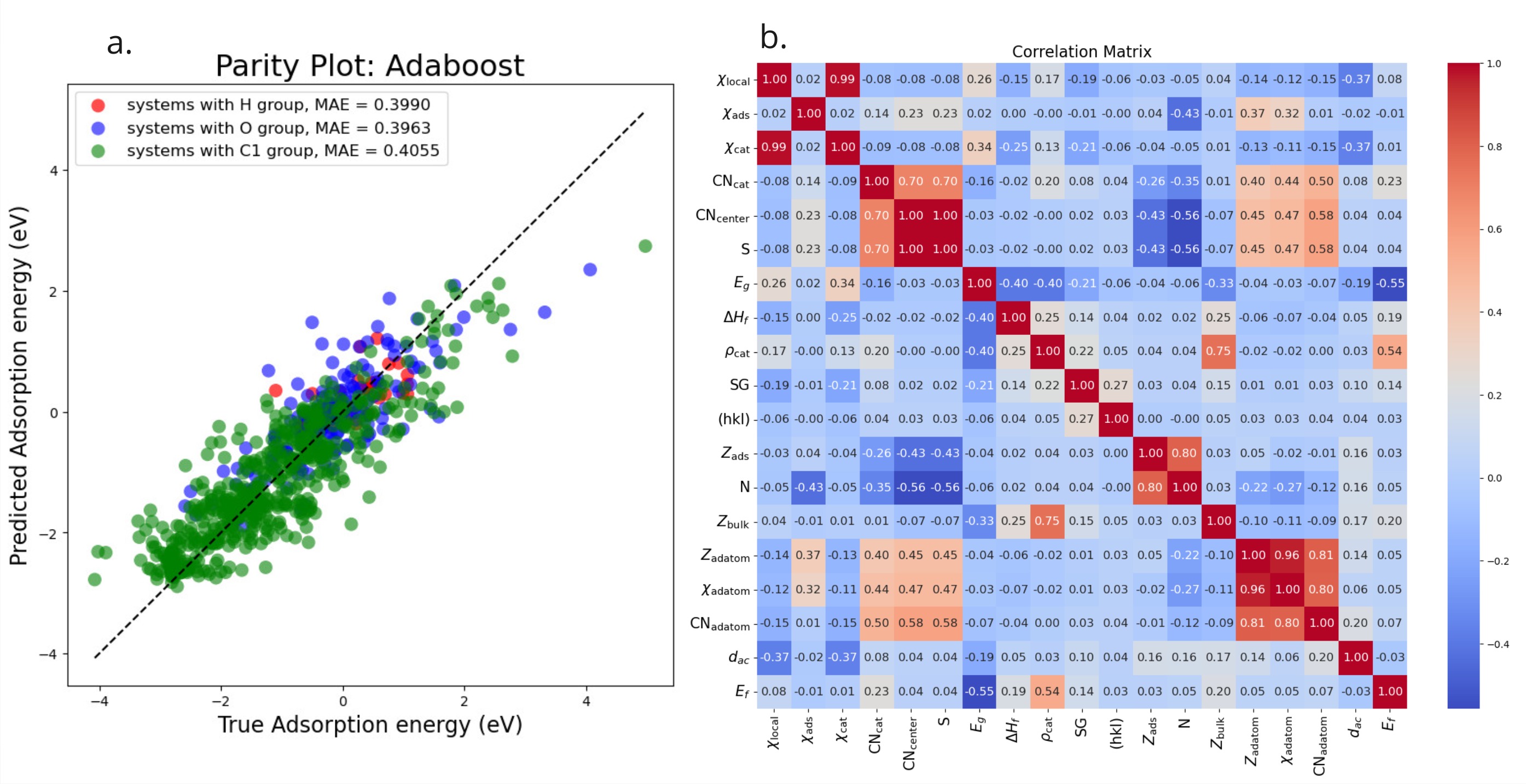} 
\caption{Performance Evaluation and Feature Correlation: (a) parity plot illustrating the performance of Adaboost Regression with Base estimator as Random Forest Regressor (Best model). MAE values are calculated for systems with H, O and C1 group. (b) Correlation matrix depicting the relationships between input features. High values of correlation coefficient have been found amongst a few features such as between Local electronegativity and Catalyst electronegativity, Site type and coordination number of adsorbate molecule. }
\label{fig:parity}
\end{figure*}

In the post-hoc analysis, the primary aspect of model explainability is carried out through SHAP analysis \cite{NIPS2017_7062}. Before conducting the SHAP analysis, a correlation matrix (Figure \ref{fig:parity}(b)) was generated to assess the relationships between input features. The matrix reveals several key relationships, such as a strong positive correlation between \( X_{\text{ads}} \) and \( X_{\text{cat}} \), indicating potential redundancy. Conversely, features like band gap \( E_g \) and effective coordination number \( CN_{\text{cat}} \) exhibit weak correlations, suggesting independent contributions to the model. Additionally, properties like \( \Delta H_f \) (enthalpy of formation) and \( \rho_{\text{cat}} \) (density of the catalyst) show moderate correlations with other features, implying they may play a multifaceted role in adsorption energy predictions. This analysis helps in eliminating highly correlated features to reduce redundancy and improve model efficiency.

The SHAP library has been widely used in various domains, including materials science, to interpret black box machine learning models \cite{lundberg2017unified}. This technique computes Shapley values for all features, thereby elucidating each feature's contribution to the prediction of adsorption energy by assessing the impact of each feature across different combinations of feature values. 

\begin{figure*}[!ht]
\centering
\includegraphics[width=0.5\textwidth]{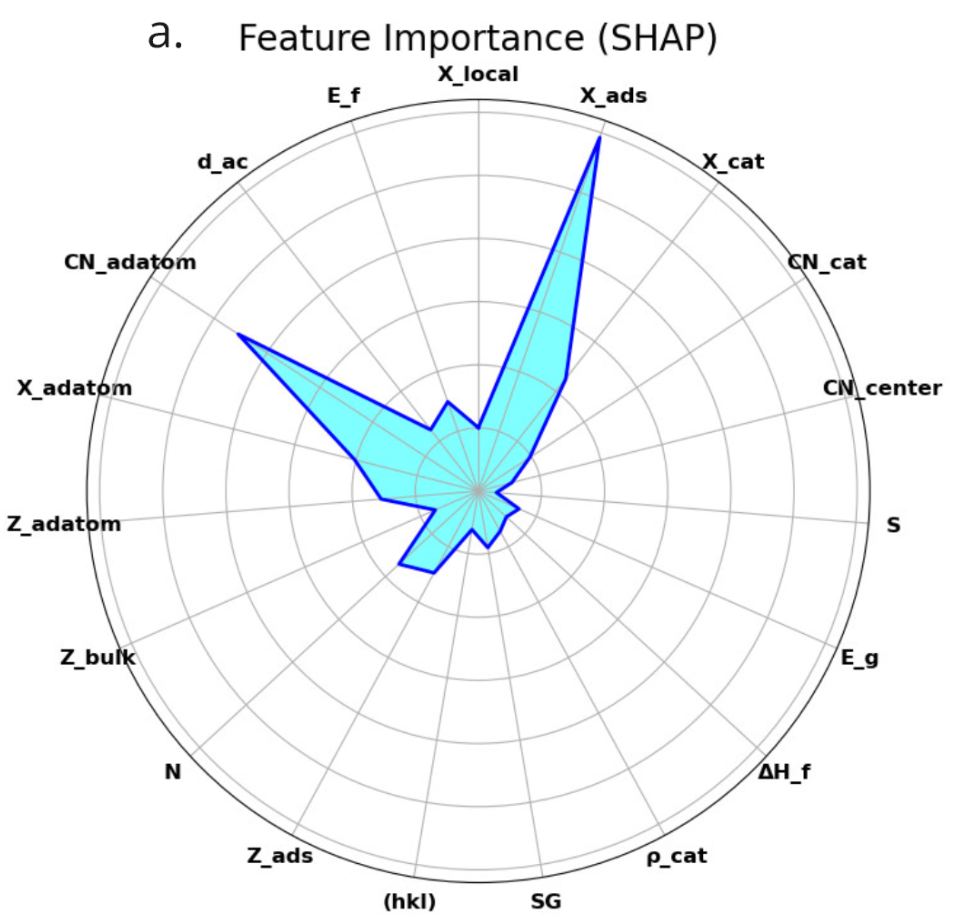}
\includegraphics[width=1.0\textwidth]
{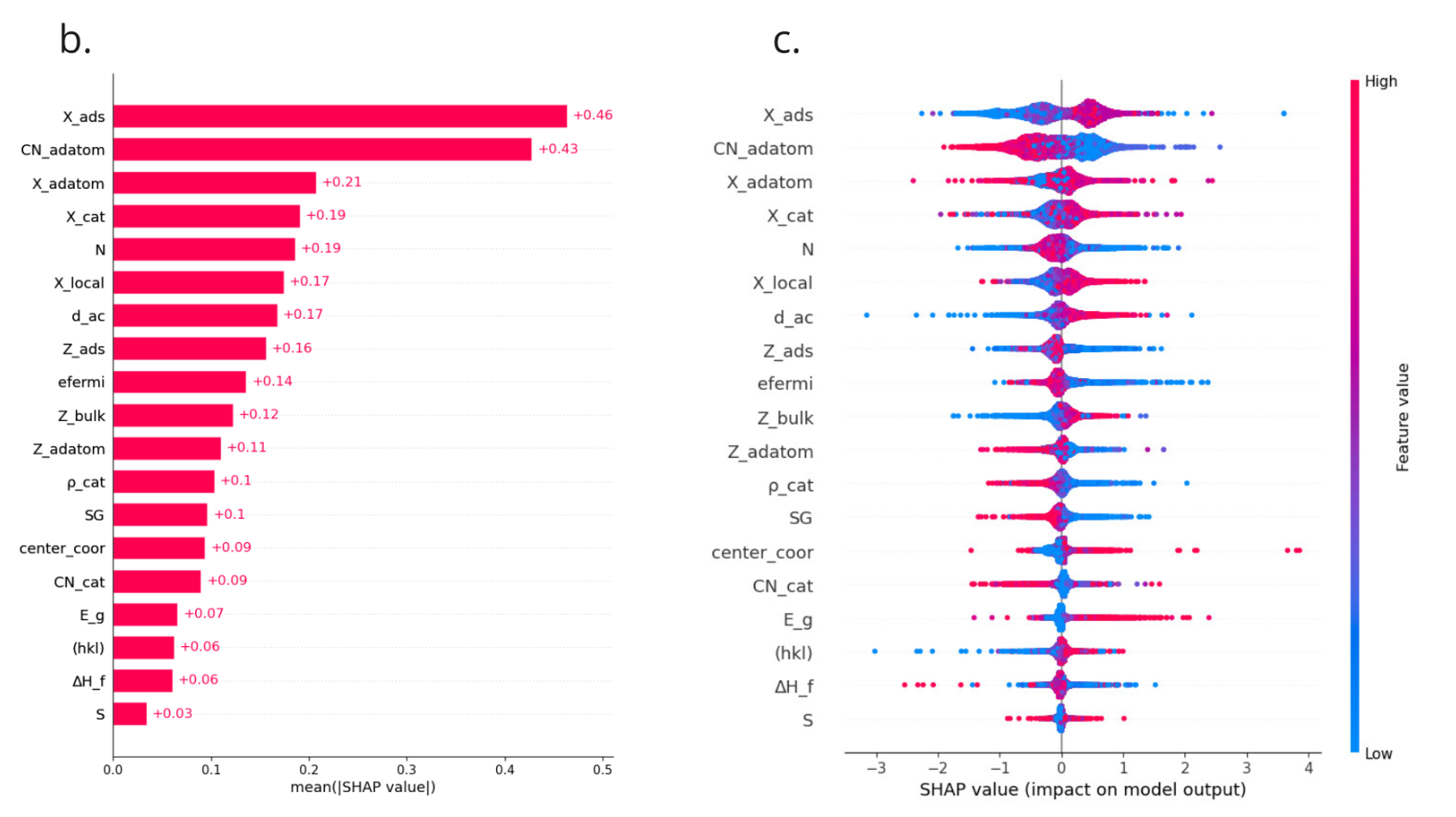} 
\caption{SHAP Analysis: (a) Radar plot illustrating feature importance based on Shapley values. (b) Summary Bar Plot presenting all features. (c) Beeswarm plot depicting the relationship between feature values and Shapley values.}
\label{fig:SHAP_analysis}
\end{figure*}

Figure \ref{fig:SHAP_analysis}(a) and Figure \ref{fig:SHAP_analysis}(b) display the feature importance using a radar chart and a summary plot, respectively. From the feature importance, the top five features (higher Shapley values) are adsorbate electronegativity (\( \chi_{\text{ads}} \)), coordination number of the adsorbing atom (\( CN_{\text{adatom}} \)), electronegativity of the adsorbing atom (\( \chi_{\text{adatom}} \)), catalyst electronegativity (\( \chi_{\text{cat}} \)), and the number of atoms in the adsorbate (\( N_{\text{ads}} \)), with Shapley values of 0.46, 0.43, 0.25, 0.19, and 0.19, respectively. Electronegativity stands out as the most significant feature among the catalyst-related factors, which is consistent with findings reported in the literature (as referenced in Table \ref{tab:shap_values}). Three of the top five features—electronegativity of the adsorbate, adsorbing atom, and catalyst surface atoms—underscore the pivotal role of electronegativity in determining adsorption behavior on catalytic surfaces.

Conversely, a few features exhibit lower importance. For instance, Formation energy \( \Delta H_f \), which plays a crucial role in single-atom catalysts as seen in Ha et al.\cite{ha2021tuning}, demonstrates limited relevance in our case, which includes heterogeneous catalysts. This may be due to its nature as a bulk property, making it less effective in predicting the interactions between adsorbates and catalytic surfaces in heterogeneous systems, as reflected in its low Shapely value (+0.06). Similarly, the band gap $E_g$, a bulk electronic property, holds minimal predictive power for surface interactions, with a Shapely value of +0.06. 

Moreover, there appears to be a stronger influence of adsorbate properties compared to catalyst properties in determining adsorption behavior. This may suggest that adsorption energy is more dependent on the type of adsorbate, possibly due to the limited variety of adsorbates, with only three categories (H, O \& C1) represented in the dataset, compared to the broader range of catalysts. This imbalance in the number of adsorbates versus catalysts could influence the observed feature importance, highlighting the significant role of data distribution in shaping the outcomes of the feature importance analysis.

Previous studies have shown varying relationships between catalyst electronegativity and adsorption energy. Gao, Wang, et al. proposed an inverse relationship specifically for transition metal catalysts, where higher electronegativity leads to lower adsorption energy \cite{gao2020determining}. Conversely, Trasatti's research suggests a direct proportionality between metal catalyst electronegativity and adsorption energy \cite{trasatti1972electronegativity}.

Our analysis reveals a positive correlation between catalyst electronegativity (\( \chi_{\text{cat}} \)) and adsorption energy. This result is evident in the beeswarm plot analysis shown in Figure \ref{fig:SHAP_analysis}(c), where significant clustering of data points and feature values for catalyst electronegativity can be observed for positive Shapley values. Unlike the findings by Gao and Wang, which show an inverse relationship between catalyst electronegativity and adsorption energy for transition metal catalysts, our dataset includes a broader range of adsorbates—hydrogen, oxygen, and carbon-based—paired with metal as well as non-metal catalysts. This diversity may account for the differences observed. This underscores the intricate nature of catalyst-adsorbate interactions, where various factors such as the composition of the dataset and the choice of adsorbates and catalysts play crucial roles in determining the observed outcomes. Additionally, a similar trend is observed for adsorbate electronegativity (\( \chi_{\text{ads}} \)) and local electronegativity (\( \chi_{\text{local}} \)), supporting our findings and emphasizing the complexity of catalyst-adsorbate interactions.

We note a denser clustering of data points with higher feature values corresponding to negative Shapley values for the number of adsorbate atoms in Figure \ref{fig:SHAP_analysis}(c). This trend suggests a diminished prediction of adsorption energy with an increased number of adsorbate atoms. This relationship may be attributed to the limited number of adsorbates used in the dataset, with a greater number of adsorbate atoms typically observed in the C1 category of adsorbates, which predominantly contributes to this relation. 


\begin{figure*}[!ht]
\centering
\includegraphics[width=0.6\textwidth]{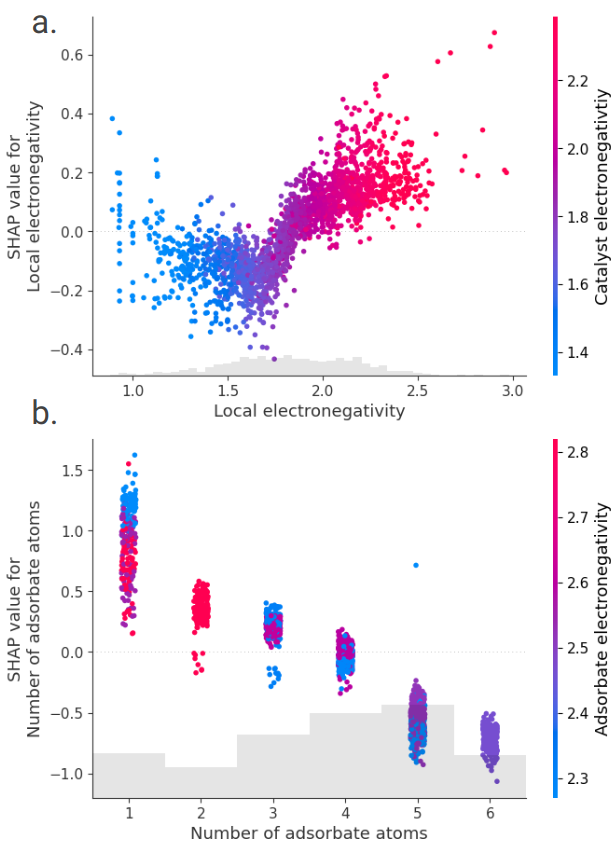} 
\caption{Feature-Feature Correlation: (a) Scatter plot illustrating the relationship between catalyst electronegativity and local electronegativity, indicating higher values of catalyst electronegativity with increasing local electronegativity. (b) Scatter plot demonstrating the correlation between the number of adsorbate atoms and adsorbate electronegativity, revealing a tendency for adsorbate electronegativity to decrease with an increase in the number of adsorbate atoms.}
\label{fig:feature_feature_rel}
\end{figure*}

In adsorption studies, surface characteristics are typically the primary focus, rather than bulk properties. However, for this dataset, the electronegativity of the catalyst bulk may still provide valuable insights into surface properties. This inference is supported by Figure \ref{fig:feature_feature_rel}, which explores feature-feature correlations and reveals two notable findings. Firstly, Figure \ref{fig:feature_feature_rel}(a) demonstrates a positive correlation between mean catalyst electronegativity ($\chi_{cat}$) and local electronegativity ($\chi_{local}$). This relationship reflects the influence of catalyst atoms within the first neighboring shell at the adsorption site. It logically follows that the electronegativity of a subset of surface atoms contributes to the overall mean electronegativity of the catalyst lattice. Additionally, a similar correlation is observed in the correlation matrix depicted in Figure \ref{fig:parity}(b), where local electronegativity and catalyst electronegativity exhibit a high correlation coefficient. This suggests a notable association between surface and bulk electronegativity in this dataset. If a strong correlation between bulk and surface characteristics is established for the target dataset, bulk characteristics could serve as proxies for surface properties, potentially offering greater flexibility in data acquisition, particularly for experimental studies.

The increase in the number of adsorbate atoms correlates with a decrease in overall adsorbate electronegativity, as depicted in Figure \ref{fig:feature_feature_rel}(b). This trend is evident from the transition in color scheme from orange (indicating high adsorbate electronegativity) to purple (indicating lower adsorbate electronegativity). The limited variety of adsorbate types in our dataset likely amplifies the role of adsorbates as key determinants in the adsorption process. The decrease in electronegativity could be attributed to the formation of chemical bonds between the adsorbate atoms. As more adsorbate atoms come into proximity, bonding interactions occur, causing a redistribution of electron density within the adsorbate system and thereby lowering the overall electronegativity \cite{cioslowski1993electron}.Adsorbate bonding and composition are key to catalyst interactions, directly affecting adsorption efficiency. This insight can guide the construction of more robust and targeted datasets for adsorption studies.

\subsection{Correlation from Symbolic Regression}

Symbolic regression is a machine learning-based regression method aimed at identifying an underlying mathematical expression that best describes the relationship between input features and output. In our study, we employed the PySR and SISSO++ libraries, which are open-source tools designed to find an interpretable symbolic expression that optimizes a specified objective \cite{cranmer2023interpretable, scheffler2022sisso++}. In our study, we selected the top five important features with the highest Shapely values from the dataset and aimed to derive meaningful mathematical expressions related to adsorption energy. After removing highly correlated features and conducting post-hoc XAI analysis, the chosen five features were: the number of adsorbate atoms (N), adsorbate electronegativity (\( \chi_{\text{ads}} \)), coordination number of the adsorbing atom (\( CN_{\text{adatom}} \)), local electronegativity (\( \chi_{\text{local}} \)), and catalyst electronegativity (\( \chi_{\text{cat}} \)). These features serve as the input for deriving interpretable, physically meaningful equations using PySR and SISSO++.

PySR generates candidate equations using genetic programming and evaluates them with a user-defined loss function (e.g., RMSE, MAE)\cite{cranmer2305interpretable}. 
SISSO++ employs a compressed-sensing-based approach to generate candidate equations by minimizing a user-defined loss function, allowing for interpretable and accurate models \cite{scheffler2022sisso++}. Both models, PySR and SISSO++, were configured with similar settings to generate equations at different complexity levels. SISSO++ outperformed PySR on both metrics. For more details on the algorithms, refer to the Methods section.

In this framework, the inputs consist of the dataset containing selected important input features and corresponding output label i.e. adsorption energy, while the output of the PySR and SISSO++ models is a set of mathematical equations that represent the relationships between the input features and output label. These equations serve as interpretable models that capture the underlying patterns and dependencies within the data.

 \begin{table}[!htbp]
\centering
\caption{Comparison of PySR equations and SISSO++ equations}
\label{tab:pysr_sisso_equations}
\renewcommand{\arraystretch}{1.5} 
\resizebox{\textwidth}{!}{%
{\normalsize 
\begin{tabular}{lll}
\hline
\textbf{PySR equations}  \\ \hline
$E_{\text{ads}} = (-0.26338 \cdot N)$\\

$E_{\text{ads}} = \frac{\chi_{\text{local}} - N}{\chi_{\text{ads}}}$\\

$E_{\text{ads}} = \left( \chi_{\text{local}} + (-0.49684 \cdot N) \right) - 0.78192$\\

$E_{\text{ads}} = \frac{\chi_{\text{ads}} - \left( N + \frac{CN_{\text{adatom}}}{\chi_{\text{local}}} \right)}{\chi_{\text{ads}}}$\\

$E_{\text{ads}} = ((\chi_{\text{ads}} - (N + (CN_{\text{adatom}} / \chi_{\text{local}}))) / (\chi_{\text{ads}}- 0.18890163))$\\
\hline
\textbf{SISSO++ equations}  \\ \hline
$E_{\text{ads}} = c_0 + a_0 \cdot \left(\frac{\chi_{\text{local}} - N}{\chi_{\text{ads}}^6}\right)$ \\
$E_{\text{ads}} = c_0 + a_0 \cdot \left(CN_{\text{adatom}} - \chi_{\text{ads}} + \left|CN_{\text{cat}} - N\right|\right) + a_1 \cdot \left(\frac{\chi_{\text{local}} - N}{\chi_{\text{ads}}^6}\right)$ \\
$E_{\text{ads}} = c_0 + a_0 \cdot \cos(\exp(\chi_{\text{ads}})) + a_1 \cdot \left(\left|CN_{\text{cat}} - N\right| + CN_{\text{adatom}}\right) + a_2 \cdot \left(\frac{\chi_{\text{local}} - N}{\chi_{\text{ads}}^3}\right)$ \\
$E_{\text{ads}} = c_0 + a_0 \cdot \cos(\chi_{\text{ads}}^2) + a_1 \cdot \left|\left(\chi_{\text{ads}} + \chi_{\text{local}}\right) - (\chi_{\text{ads}} \cdot \chi_{\text{local}})\right| + a_2 \cdot \left(CN_{\text{adatom}} - \chi_{\text{ads}} + \left|CN_{\text{cat}} - N\right|\right) + a_3 \cdot \left(\frac{\chi_{\text{local}} - N}{\chi_{\text{ads}}^6}\right)$ \\ 
$E_{\text{ads}} = c_0 + a_0 \cdot \left|\chi_{\text{ads}} - \chi_{\text{local}} - \exp(-CN_{\text{adatom}})\right| + a_1 \cdot \sin(\chi_{\text{local}})^6 + a_2 \cdot \cos(\chi_{\text{ads}}^2) + a_3 \cdot \left(\left|CN_{\text{cat}} - N\right| + CN_{\text{adatom}}\right) + a_4 \cdot \left(\frac{N}{\chi_{\text{ads}} \cdot \chi_{\text{ads}}^3}\right)$ \\
\hline
\end{tabular}%
}}
\end{table}

The equations generated by PySR and SISSO++, as shown in Table \ref{tab:pysr_sisso_equations}, consist of five equations for each method, presented in increasing levels of mathematical complexity. PySR equations are simpler, utilizing basic arithmetic operations, while SISSO++ produces more complex equations with higher-order terms that capture subtle, non-linear relationships in the data. PySR begins with a single input feature at complexity 1 and gradually includes more features as complexity increases, integrating all key features by complexity 5 (Table \ref{tab:pysr_sisso_equations}). In contrast, SISSO++ uses a broader range of parameters from the outset, incorporating three features at complexity 1 and utilizing all features by complexity 3. This leads to faster convergence rates and enhances its applicability in computational modeling. Another notable advantage of SISSO++ is its ability to consider feature units, ensuring physical relevance and interpretability of the resulting equations. For this study, feature units were specified during the SISSO++ setup, providing context that PySR lacks.

Although SISSO++ and PySR exhibit relatively high root mean square error (RMSE) values of 0.8 eV and 1.2 eV, respectively (details in the SI), SISSO++ successfully captures the correlation between input features and adsorption energy, aligning with the findings of Gao et al.\cite{gao2020determining}. Gao’s work shows a direct proportionality between adsorption energy and the effective coordination number, as seen in Equation \ref{eq:gao_equation}. The first term contains $\psi$ which is electronic descriptor, $CN_{cat}$ in second term is effective coordination number and $\alpha$, $\lambda$ and $\theta$ are system-specific parameters. Similarly, SISSO++ identifies this proportionality from the second dimension onward, where $E_{\text{ads}}$ is directly proportional to $|CN_{cat}|$, the effective coordination number (as seen in table \ref{tab:pysr_sisso_equations}). These results suggest that at appropriate complexity levels, SISSO++ effectively captures critical feature interactions and underlying relationships, improving predictive accuracy while maintaining alignment with established literature.

\begin{equation}
 E_{\text{ads}} = 0.1 \cdot \alpha \cdot \psi + \lambda \cdot CN_{cat} + \theta
\label{eq:gao_equation}
\end{equation}

In addition to capturing important feature expressions, the SISSO++ library offers a deterministic approach to model discovery, ensuring reproducible results—an advantage over the genetic programming methods like  PySR which are stochastic in nature. SISSO++ also supports multi-task learning, allowing it to solve multiple related tasks simultaneously and leverage shared information to improve performance across all tasks.

\section{Conclusion}

Our study successfully integrates explainability into machine learning models through post-hoc SHAP analysis and symbolic regression, highlighting the importance and correlations of key features. By merging machine learning techniques with XAI, we bridge the gap between data-driven approaches and domain-specific knowledge, offering insights into the intricate relationships between input structural features and adsorption energy. One key finding is the direct proportionality between effective coordination number and adsorption energy in catalytic systems. This is corroborated by multiple analytical approaches: SHAP analysis identified effective coordination number as one of the top five influential features, while symbolic regression method, particularly SISSO++, expressed this direct relationship in its generated equations. Notably, these findings align with and further support the research of Gao et al., underscoring the robustness of this relationship across methodologies.

Additional insights from feature-feature correlations via SHAP analysis revealed a strong correlation between catalyst and local electronegativity, and how adsorbate electronegativity decreases with increase in number of adsorbate atoms for the dataset. The XAI techniques provide crucial insights into ML model predictions by highlighting feature importance and the impact of feature values on adsorption energy. This facilitates focused attention on relevant features, enabling symbolic regression to derive better relationships that can be leveraged for future catalyst optimization. In summary, our study establishes a robust framework that integrates machine learning techniques with XAI, using extensive datasets such as OC20 to improve catalyst design through enhanced model explainability.

\section{Methods}

\subsection{Dataset and Feature selection}

The dataset utilized in this study is sourced from the OC20 dataset, which comprises 872,000 trajectories obtained from DFT calculation results, making it suitable for training ML models. From these trajectories, we extracted final frames to obtain relaxed structures of adsorbate-catalyst systems. The OC20 dataset encompasses catalyst systems featuring a diverse array of reactive elements, spanning nonmetals, alkali metals, alkaline earth metals, metalloids, transition metals, and post-transition metals. In contrast, the adsorbate systems consist of 82 molecules categorized into oxygen or hydrogen, C1 molecules, C2 molecules, and nitrogen-containing molecules.

For this study, we curated a dataset for post-hoc analysis by extracting a subset of the OC20 dataset, specifically focusing on systems containing oxygen (O), hydrogen (H), and C1 category adsorbates. This focus on small molecules minimizes the impact of interatomic interactions within the adsorbate molecules, simplifying comparisons with previous theory-based studies. Following this selection process, the dataset was narrowed down to a subset consisting of 24503 systems. To extract features from this structure data, ASE and Pymatgen packages are utilized, enabling the extraction of relevant information from the relaxed structures of each trajectory. The adsorption energy is retrieved from the OC20 dataset. Ultimately, the structured data, comprising 19 input features and adsorption energy as the output label, is fed into the ML models.

\subsection{SHAP approach}

We use SHAP library to perform post-hoc XAI analysis. SHAP is a technique based on game theory that helps explain how machine learning models make decisions. It connects fair credit distribution with simple explanations by using classic Shapley values. Shapely values are calculated using the Shapley regression values, a concept from cooperative game theory. Shapley regression values are feature importances for linear models in the presence of multicollinearity. This method requires retraining the model on all feature subsets $S \subseteq F$. Mathematically, the Shapley value for a feature \( i \) is calculated as follows:
\begin{equation}
\label{eq:SHAP_eqn}
\phi_i = \sum_{S \subseteq F \setminus \{i\}} \frac{|S|!(F-|S|-1)!}{|F|!} [f_{S \cup \{i\}}(x_{S \cup \{i\}}) - f_S(x_S)
\end{equation}

where \( F \) is the total number of features, \( S \) is a subset of features excluding feature \( i \), \( x_S \) represents the input data with features in subset \( S \), and \( x_{S \cup \{i\}} \) represents the input data with features in subset \( S \) along with feature \( i \). \( f_{S \cup \{i\}} \) is the model prediction using features in subset \( S \) along with feature \( i \) where as  \( f_S\) is the model prediction with feature \( i \) withheld. The term \( |S|!(M-|S|-1)!/M! \) represents the probability of including feature \( i \) in a subset \( S \) of size \( |S| \), which is adjusted by the number of permutations of subsets containing \( i \).

In essence, the Shapley value for a feature represents the average change in the model prediction when including that feature compared to all possible combinations of features, weighted by their probabilities. Positive Shapley values signify a positive correlation between features, indicating that as one feature increases, the other feature tends to increase as well. Conversely, negative Shapley values denote a negative correlation, implying that as one feature increases, the other tends to decrease. Our study visualizes Shapley values of all features in various charts such as summary plot, beeswarm plot and scatter plots for a comprehensive understanding of feature importances and correlations.

\subsection{PySR}

PySR (Symbolic Regression using Python) is an open-source library to perform symbolic regression on the dataset \cite{cranmer2305interpretable}. The core of PySR model's algorithm involves iteratively evolving populations of mathematical expressions through tournament selection \cite{goldberg1991comparative}, mutation, crossover, simplification, and optimization \cite{broyden1970convergence, mogensen2018optim}. These populations evolve independently, with occasional migration of individuals between them. The evolve-simplify-optimize loop refines the discovered equations by simplifying them to equivalent forms and optimizing constants. This iterative process continues until satisfactory equations representing the relationships between input features and output labels are discovered. 

The mathematical equations are produced through a process of evolutionary optimization, where populations of mathematical expressions are iteratively evolved and refined to best fit the dataset. This involves the generation of candidate expressions, their evaluation based on predefined fitness criteria (e.g., goodness of fit to the data), and the application of genetic operators (mutation, crossover) to create new expressions. Additionally, simplification techniques are employed to reduce the complexity of the expressions while preserving their accuracy, and optimization methods are applied to fine-tune the constants within the equations. Through this iterative process, the framework systematically explores the space of mathematical expressions to discover equations that effectively capture the relationships between the input features and output label.

To further refine these equations and mitigate overfitting, we imposed constraints during symbolic regression. We adopted RMSE as the loss function to enable direct comparison with the SISSO++ library at different complexity levels. Given the large dataset, we optimized the model with 50 iterations and enabled Turbo mode, which uses the Loop Vectorization method to accelerate the evaluation among a pool of equations. The \texttt{max\_size} parameter was set to 7, limiting equation complexity to 7, while the \texttt{max\_depth} was set to 5, restricting the maximum depth of the mathematical expressions or trees. We set lower values for \texttt{max\_depth} and \texttt{max\_size} to prioritize simplicity and interpretability, ensuring that the generated equations remain easy to understand and less prone to overfitting. Additionally, we enabled the \texttt{denoise} option, allowing the model to filter out noise in the input data and focus on the underlying relation rather than fitting to noise.

\subsection{SISSO++}

SISSO++ is an advanced feature selection and model construction algorithm that combines two key methods: Sure Independence Screening (SIS) and Subspace Optimization (SO). The SIS method, crucial for handling high-dimensional problems, begins with a large pool of potential features or descriptors. It calculates the correlation between each feature and the target variable (in this case, adsorption energy), then ranks and screens features based on these correlations. A subset of the most promising candidates is selected, and an independence check is performed to avoid redundancy and multicollinearity. This approach effectively reduces the dimensionality of the problem by retaining only the most relevant and independent features. Following the SIS step, the SO method comes into play. It performs a combinatorial optimization within the subspace of selected features to identify the best descriptor (or combination of descriptors) for the target property. SISSO++ applies this SIS-SO process iteratively, refining the selected features through multiple rungs.

The SISSO++ model settings used in this work are optimized to match the PySR model configuration for direct comparison. A descriptor dimensionality of 5 is chosen to capture sufficient information while controlling model complexity. A descriptor dimensionality of 5 (\texttt{desc\_dim = 5}) is chosen to capture sufficient information while controlling model complexity. In each iteration, 3 features are selected (\texttt{n\_sis\_select = 3}) to ensure the most relevant variables are prioritized. The model progresses through 2 rungs (\texttt{max\_rung = 2}), refining the selected features iteratively. To account for residual errors, 1 additional descriptor is added (\texttt{n\_residual = 1}). The dataset used is specified by \texttt{data.csv}, and a 20\% fraction of the data is reserved for testing purposes (\texttt{leave\_out\_frac = 0.2}) to ensure robust model validation similar to PySR model. Only the top-performing model is stored (\texttt{n\_models\_store = 1}), and the intercept is not fixed (\texttt{fix\_intercept = false}) to allow for more flexible linear fitting. These settings are selected to achieve a balance between model performance and computational efficiency.

\section{Data and Software Availability statement}

Both the Python code for data retrieval and the data employed in this study are available on GitHub at the following link: \url{https://github.com/tirtha-v/eXplainable_AI}.









\bibliography{main}

\end{document}